\documentclass{l4dc2027}
\makeatletter
\def\ps@jmlrtps{\let\@mkboth\@gobbletwo\def\@oddhead{\scriptsize Submitted to the 9th Annual Learning for Dynamics and Control Conference (L4DC 2027)\hfill}\let\@evenhead\@oddhead\def\@oddfoot{\@titlefoot}\let\@evenfoot\@oddfoot}
\makeatother

\usepackage{times}
\usepackage{booktabs}

\newcommand{\Law}{\operatorname{Law}}

\newcommand{\E}{\mathbb{E}}
\newcommand{\R}{\mathbb{R}}
\newcommand{\epsC}{\epsilon_C}

\title[Lift--project SDEs for snapshot data]{Learning hidden-state dynamics from unpaired snapshots\\ with lift--project SDEs}

\ifanonsubmission
\author{\Name{Anonymous Authors}\\ \addr Anonymous Institution}
\else
\coltauthor{%
 \Name{Qi Wang} \Email{qwang4@sdsu.edu}\\
 \addr Department of Aerospace Engineering, San Diego State University, San Diego, CA, USA
 \AND
 \Name{Uduak George} \Email{ugeorge@sdsu.edu}\\
 \addr Department of Mathematics and Statistics, San Diego State University, San Diego, CA, USA
 \AND
 \Name{Gustaaf Jacobs} \Email{gjacobs@sdsu.edu}\\
 \addr Department of Aerospace Engineering, San Diego State University, San Diego, CA, USA%
}
\fi

\begin{document}

\maketitle

\begin{abstract}%
Many systems are observed only through unpaired snapshots, with population samples available at each time point but no correspondence between samples at different times.
These snapshots determine the one-time distributions but leave the temporal coupling unspecified, requiring a modeling prior to connect successive times.
We propose the lift--project stochastic differential equation (SDE), an autonomous SDE on the observed coordinate and a few hidden coordinates, trained only to match the snapshots.
Most existing methods, including time-dependent SDEs, Schr\"odinger bridges and optimal transport, evolve the observed coordinate alone.
For a scalar observation, the classical comparison theorem implies that all such models have non-negative lagged covariance, so on data whose lagged covariance turns negative, their error has a computable floor.
The lift--project SDE is not bound by this floor.
On two representative partially observed systems, the Duffing and pendulum oscillators, at comparable fit to the snapshots, the proposed method reduces the lagged-covariance error by factors of $2.4$ and $6.6$ relative to the strongest baseline and is the only model below the error floor; one hidden coordinate is already enough for such improvement.
Furthermore, if a second experiment changes only the initial state of the hidden coordinate, no single model that evolves the observed coordinate alone can fit both experiments, whereas a single lift--project SDE can.
\end{abstract}


\section{Introduction}
\begin{figure}[t]
\centering
\includegraphics[width=\textwidth]{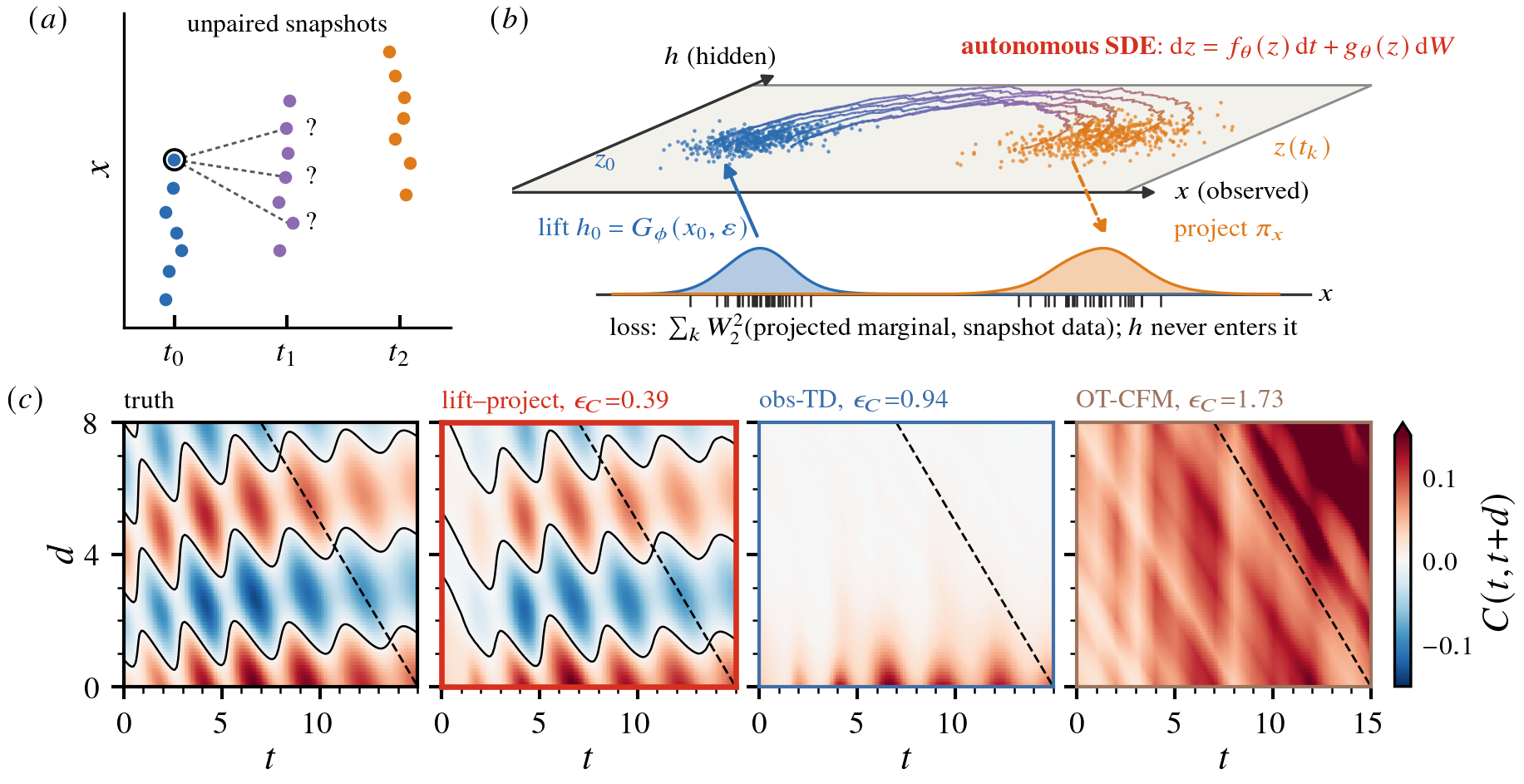}
\caption{The lift--project SDE recovers the temporal coupling of a partially observed system.
(a) Unpaired snapshots: samples at different times cannot be paired.
(b) Schematic of the method: lift the observable $x$ to $z=(x,h)$ and advance with an autonomous SDE, followed by a projection back to $x$.
(c) Lagged covariance $C(t,t+d)$ of the true Duffing process and of three models learned only from its snapshots, with their errors $\epsC$.}
\label{fig:problem}
\end{figure}

Many systems can be observed only through time-indexed marginal distributions rather than individual trajectories. Consider measuring gene expression in a tissue at daily intervals. Because sequencing destroys each cell, each measurement consists of a fresh population sample: no cell can be observed at more than one time point.
Learning dynamics from such time-course data is a central task in developmental biology \citep{schwartz2021yolk,schwartz2023dynamic}.
Particle tracking at high density, where velocities are not measured \citep{dominguez2021lagrangian,dominguez2023sparser}, and ensemble weather forecasts give the same kind of data.
In each case, the data are distributions $p_{t_1},\dots,p_{t_K}$ of an observed coordinate $x$, and the correspondence between samples at different times is absent (Figure~\ref{fig:problem}a).

To learn dynamics from these snapshots, a method has to decide how the mass at time $t_k$ moves to time $t_{k+1}$.
The temporal coupling determines two-time statistics, such as lagged covariances, as well as path-dependent quantities, such as first-passage times. Yet the snapshots alone determine only the one-time marginals and leave this temporal structure unspecified \citep{weinreb2018fundamental,lavenant2024toward}.
The coupling therefore comes from the method's \emph{modeling prior}: its architecture, its way of pairing samples across times, and its training objective.
Existing methods choose this prior in different ways.
Waddington-OT pairs consecutive snapshots by entropic optimal transport \citep{schiebinger2019optimal}. In the appropriate continuum limit, this construction is provably equivalent to a diffusion with gradient-form drift \citep{lavenant2024toward,chizat2022trajectory}.
TrajectoryNet and MIOFlow regularize continuous normalizing flows toward transport-based or low-dimensional manifold structure \citep{tong2020trajectorynet,huguet2022mioflow}, flow and bridge matching avoid simulation during training \citep{tong2024otcfm,tong2024sf2m}, multi-marginal flow matching shares parameters across times and conditions \citep{rohbeck2025modeling}, and neural SDEs can be fitted to population losses \citep{hashimoto2016learning,li2020scalable}.
These \emph{observed-state models} evolve the observed coordinate alone through a time-dependent drift or a transport map between snapshots.

This restriction is consequential, because in physical systems the observed coordinate is often only part of the state: a measurement may record position but not velocity, or concentration but not the flow that transports it \citep{wang2022observable,wang2025domain,dominguez2025inference}.
The projection of a Markovian system onto an observed coordinate need not remain Markovian, and its memory shapes temporal statistics such as the lagged covariance, which an observed-state model may then fail to reproduce.
Hidden coordinates are routine when trajectories are measured over time: neural SDEs \citep{li2020scalable,kidger2021neural,kidger2022thesis} and latent ODEs \citep{rubanova2019latent} learn them from trajectories, Koopman lifting \citep{lusch2018deep,korda2018linear,l4dc_koopman} and Mori--Zwanzig models \citep{mori1965transport,zwanzig2001nonequilibrium,lin2021datadriven} add coordinates to make the dynamics simpler, and data assimilation recovers unobserved forcing, sources or states from time-resolved measurements \citep{wang2019spatial,mons2019kriging,wang2026mitigating}.
With snapshots only, \citet{wang2026characteristic} learn hidden ODE dynamics along the characteristics of a Liouville equation in an augmented state space, and Liouville models of particle-laden flow evolve joint densities driven by a few random variables \citep{dominguez2024highorder}.

We propose the \emph{lift--project SDE}, which lifts the observed coordinate with a few hidden coordinates, evolves the lifted state with one autonomous SDE, and projects back to the observed coordinate.
It is trained on unpaired snapshots alone.
For a scalar observation, the classical comparison theorem implies that every observed-state diffusion, including time-dependent SDEs, Schr\"odinger bridges and transport flows, has non-negative lagged covariance.
From this we derive a computable error floor on data whose lagged covariance turns negative, and the lift--project SDE is not bound by it.
On partially observed Duffing and pendulum oscillators, at comparable fit to the snapshots, the lift--project SDE reduces the lagged-covariance error several-fold relative to the strongest baseline, with a single hidden coordinate.
If two experiments share the same dynamics but differ only in their initial hidden states, no single model evolving the observed coordinate alone can fit both experiments, but a single lift--project SDE can.

\section{Problem setup}
\label{sec:setting}

\begin{figure}[t]
\centering
\includegraphics[width=\textwidth]{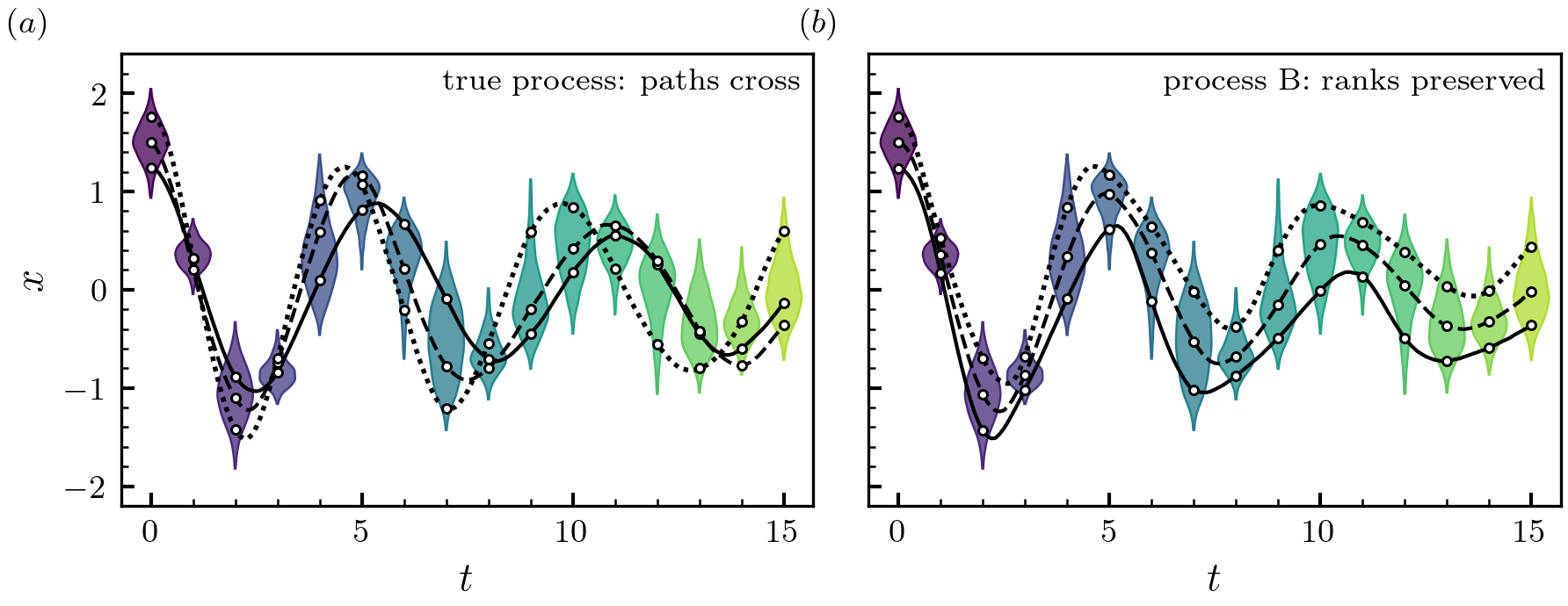}
\caption{The same marginals admit different temporal couplings.
Colored patches show the marginals $p_{t_k}$ of the Duffing position $x$ at the training times, scaled to equal width.
Lines mark three paths of $x$ in each process.
(a) The true process; paths cross because the hidden velocity carries them.
(b) The rank-preserving process B; paths never cross, yet its marginals are identical to those of the true process.}
\label{fig:snapshots}
\end{figure}

Let $Z_t\in\R^{n}$ be the full state of a system and $x_t=\pi_x(Z_t)\in\R^{d_x}$ the observed coordinate, where $\pi_x$ extracts the coordinate $x$ from $Z=(x,h)$ (the \emph{projection}).
Each experiment specifies an initial law $\mu$ of $Z_0$.
We call $p_t=\Law(x_t)$ the \emph{marginal} at time $t$.
The data are \emph{snapshots}: at each time $t_k$, $k=0,\dots,K-1$, the snapshot $\hat p_k$ consists of $N$ fresh samples of $p_{t_k}$, drawn independently of the samples at other times.
Two processes are \emph{marginal-equivalent} if they have the same $p_t$ at every $t$, so that no snapshot data can tell them apart.
Figure~\ref{fig:snapshots} shows two such processes for the Duffing oscillator of Section~\ref{sec:exp}.
One is the true process: its hidden velocity carries particles from high to low $x$ and back, so its paths cross.
The other, \emph{process~B}, keeps every particle at the same rank, $x_t=Q_t(U)$ with $Q_t$ the quantile function of $p_t$ and one uniform random variable $U$; its paths never cross, and its coupling is that of one-dimensional monotone optimal transport.
The two processes have the same marginals but different temporal couplings.

A model is defined as a stochastic process $\hat x_t$ that we can sample.
We measure its marginal fit by the quadratic Wasserstein distance $W_2$ \citep{villani2009optimal} between the model and data distributions at each time; for one-dimensional problems (scalar $x$), it is computed exactly by sorting.
$\overline W_2^{\,\mathrm{train}}$ is its mean over the training times after $t=0$, and the \emph{noise floor} is the $W_2$ between two independent samples of the same process.
We measure the temporal coupling by the lagged covariance $C(t,t+d)=\operatorname{Cov}(x_t,x_{t+d})$ on the grid $t\in\{0,1,\dots\}$, $t+d\le T$, and by its relative error
\begin{equation}
\epsC=\|\hat C-C^{\star}\|_F/\|C^{\star}\|_F ,
\end{equation}
where $C^{\star}$ is the lagged covariance of a reference process and $\|\cdot\|_F$ is the Frobenius norm; unless stated otherwise the reference is the true process, and $\epsC^{B}$ uses process~B instead.

\section{The lift--project SDE and the baselines}
\label{sec:model}

\paragraph{Lift--project SDE.}
The state is $z=(x,h)$ with hidden coordinates $h\in\R^{d_h}$, and it evolves by one autonomous SDE, $\mathrm{d}z=f_\theta(z)\,\mathrm{d}t+g_\theta(z)\,\mathrm{d}W$, with time-independent dynamics (Figure~\ref{fig:problem}b).
The lift is a generator that draws $h_0=G_\phi(x_0,\varepsilon)$ with $\varepsilon\sim\mathcal N(0,I)$, so it learns the conditional law $\Law(h_0\mid x_0)$.
The drift, the diffusion and the generator are multilayer perceptrons (MLPs).
The lifted state is then evolved by the lifted SDE. Numerically, it is integrated with Euler--Maruyama.
After the projection by $\pi_x$, the distribution of $x$ can be compared with the snapshots.
The loss never sees the latent variable $h$, and the hidden coordinates are learned only through their effect on the snapshots.
A stability prior bounds the diffusion and gives the drift a linear confining tail, so simulated paths cannot escape (compare \citealp{manek2019learning,heide2026datadriven}).
We use $d_h=2$ unless stated otherwise; with $d_h=0$ the model is a time-homogeneous SDE in $x$, which we call obs-TH (observed state, time-homogeneous).

\paragraph{Training.}
All parameters appearing in the drift, the diffusion term, and the generator are trained by minimizing the following loss function,
\begin{equation}
\mathcal L(\theta)=\sum_{k=1}^{K-1}W_2^2\big((\pi_x)_\#\,p_\theta(t_k),\,\hat p_k\big),
\label{eq:loss}
\end{equation}
where $p_\theta(t_k)$ is the law of $z$ at $t_k$ under the learned SDE and initial law, and $\hat p_k$ is the snapshot at $t_k$.
For scalar $x$, $W_2^2$ is the mean squared difference between the sorted model samples and the sorted data samples.
Within an iteration, the generator noise and Brownian increments are fixed, so each path is differentiable with respect to the parameters $\theta$, and we backpropagate through the full rollout \citep{li2020scalable,kidger2022thesis}.
We use Adam with fresh noise and data minibatches and grow the rollout horizon (details shown in Appendix~\ref{app:details}).
Unlike objectives based on optimal transport or flow matching, this loss does not require any temporal correspondence between samples in different snapshots.
Algorithm~\ref{alg:train} summarizes the training loop.

\begin{algorithm2e}[t]
\caption{Training algorithm of the lift--project SDE}
\label{alg:train}
\KwIn{snapshots $\hat p_k=\{x_k^{(i)}\}_{i=1}^{N}$ at $t_k$, $k=0,\dots,K-1$; hidden dimension $d_h$; step $\Delta t$; iterations $M$; batch $B$; learning rates $\eta_0>\eta_1$; clip norm $c$; initial horizon $H_0$}
\KwOut{drift $f_\theta$, diffusion $g_\theta$ and generator $G_\phi$}
initialize $f_\theta$, $g_\theta$ and $G_\phi$ as MLPs, with the stability prior applied to $f_\theta$ and $g_\theta$\;
\For{$m=0,\dots,M-1$}{
  $\eta\leftarrow\eta_1+\tfrac12(\eta_0-\eta_1)(1+\cos(\pi m/M))$; \quad $H\leftarrow\min\{T,\,H_0+(T-H_0)\,m/(M/2)\}$; \quad $k_{\max}\leftarrow$ largest $k$ with $t_k\le H$\;
  draw $B$ samples $x_0^{(b)}$ from $\hat p_0$ and $\varepsilon^{(b)}\sim\mathcal N(0,I_{d_h})$; set $z_0^{(b)}=(x_0^{(b)},G_\phi(x_0^{(b)},\varepsilon^{(b)}))$\;
  \For{$n=0,\dots,t_{k_{\max}}/\Delta t-1$}{
    draw $\xi_n^{(b)}\sim\mathcal N(0,I_{1+d_h})$; $z_{n+1}^{(b)}=z_n^{(b)}+f_\theta(z_n^{(b)})\Delta t+g_\theta(z_n^{(b)})\odot\xi_n^{(b)}\sqrt{\Delta t}$\;
  }
  \For{$k=1,\dots,k_{\max}$}{
    draw a fresh minibatch $\{y_k^{(b)}\}_{b=1}^{B}$ from $\hat p_k$ without replacement; project $\hat x_k^{(b)}=\pi_x(z^{(b)}_{t_k/\Delta t})$\;
    sort both samples into order statistics $\hat x_{k,(1)}\le\dots\le\hat x_{k,(B)}$ and $y_{k,(1)}\le\dots\le y_{k,(B)}$\;
  }
  $\mathcal L\leftarrow\sum_{k=1}^{k_{\max}}\frac1B\sum_{b=1}^{B}\big(\hat x_{k,(b)}-y_{k,(b)}\big)^2$\;
  compute $\nabla_{\theta,\phi}\mathcal L$ by backpropagation through the whole rollout, with $\varepsilon$ and $\xi$ held fixed\;
  clip the gradient to global norm $c$; take one Adam step with learning rate $\eta$\;
}
\textbf{Notation:} $u_{(b)}$ is the $b$-th smallest of $B$ values $u^{(1)},\dots,u^{(B)}$; sorting gives the exact one-dimensional $W_2^2$ between the two empirical laws.
\end{algorithm2e}

\paragraph{Baselines.}
All baselines are \emph{observed-state models}: they evolve the observed coordinate $x$ directly and contain no hidden coordinates. Besides obs-TH, we compare three types of models.
The first is obs-TD (observed state, time-dependent), an SDE for $x$ with a time-dependent drift and diffusion. By Gy\"ongy's mimicking theorem \citep{gyongy1986mimicking}, it can in principle match all marginals exactly. We use Fourier features in time and use multiple shooting for the Duffing oscillator.
The second is OT-CFM, which uses conditional flow matching with the one-dimensional optimal-transport coupling between each pair of consecutive snapshots \citep{tong2024otcfm}.
The third is SB, a Schr\"odinger-bridge model \citep{leonard2014survey,tong2024sf2m}. It uses entropically regularized transport to couple consecutive snapshots and learns the corresponding drift through Brownian-bridge matching. For the pendulum, we also consider obs-TD-warm, which initializes obs-TD with the velocity field learned by OT-CFM.
All models use the same stability prior for their drift fields, and their parameter counts agree within \(\pm 10\%\).

\section{Theory}
\label{sec:theory}

The following classical construction shows that snapshots do not determine the temporal coupling; all proofs are in Appendix~\ref{app:proofs}.

\begin{lemma}[Comonotone construction]
\label{prop:nonid}
Let $(X_t)_{t\in[0,T]}$ be a real-valued process with finite second moments whose marginal cumulative distribution functions $F_t(x)=\mathbb P(X_t\le x)$ are continuous, and let $Q_t=F_t^{-1}$ be the quantile functions.
Define $U=F_0(X_0)$ and the rank-preserving process $Y_t=Q_t(U)$.
Then (i) $\Law(Y_t)=\Law(X_t)$ for every $t$; (ii) for every pair $s<t$, $\operatorname{Cov}(Y_s,Y_t)\ge \operatorname{Cov}(X_s,X_t)$; and (iii) equality in (ii) holds if and only if the joint law of $(X_s,X_t)$ equals the comonotone coupling $\Law(Q_s(U),Q_t(U))$, that is, the ranks of $X$ are preserved between $s$ and $t$.
\end{lemma}

Pairing equal ranks across times therefore reproduces every marginal, and any process whose particles change rank has a marginal-equivalent twin with a different lagged covariance.
With $X$ the true process, $Y$ is process~B.
Gy\"ongy's mimicking theorem gives another twin: a diffusion that is Markovian in $x$ and has the same marginals as any It\^o process with bounded, nondegenerate coefficients \citep{gyongy1986mimicking}.

\paragraph{Finite-dimensional Markovian realization of exponential memory.}

Realization theory asks under what condition an input--output map can be produced by a finite-dimensional state-space system \citep{kalman1963mathematical,ho1966effective}, and stochastic realization theory asks the same question for random processes \citep{lindquist2015linear}.
For physical systems with memory, the Markovian embedding of a generalized Langevin equation gives such a realization \citep{pavliotis2014stochastic}.

\begin{lemma}[Exponential memory with colored noise]
\label{prop:gle}
Consider the generalized Langevin equation (GLE)
\begin{equation}
\dot x=v,\qquad \dot v=-\nabla U(x)-\int_0^t \mathcal K(t-s)\,v(s)\,\mathrm{d}s+\xi(t),
\label{eq:gle}
\end{equation}
with potential $U$ with Lipschitz gradient, memory kernel $\mathcal K(t)=\sum_{i=1}^m c_i^2 e^{-\lambda_i t}$, $\lambda_i>0$, and a stationary centered, colored Gaussian noise $\xi$, independent of $(x_0,v_0)$, with $\E[\xi(t)\xi(s)]=\varsigma^2\mathcal K(|t-s|)$ for a noise level $\varsigma>0$.
Let $u_1,\dots,u_m$ solve
\begin{equation}
\mathrm{d}u_i=(-\lambda_i u_i-c_i v)\,\mathrm{d}t+\varsigma\sqrt{2\lambda_i}\,\mathrm{d}W_i
\label{eq:aux}
\end{equation}
with independent Brownian motions $W_i$, and set $\dot v=-\nabla U(x)+\sum_i c_i u_i$.
Then $(x,v,u_1,\dots,u_m)$ is an autonomous Markovian SDE.
If the auxiliary variables start from their stationary law $u_i(0)\sim\mathcal N(0,\varsigma^2)$, independent of each other, of $(x_0,v_0)$ and of the $W_i$, then its $x$-component has the same law as the solution of \eqref{eq:gle}.
\end{lemma}

Each exponential memory mode thus adds one state variable; the lift--project SDE uses this structure, with hidden coordinates learned from data rather than fixed by physics.

\paragraph{Limitations of observed-state models.}
\label{sec:assoc}

The next classical result separates observed-state models from the lift--project SDE.

\begin{proposition}[Positive temporal association]
\label{prop:pos}
Let $b(t,x)$ and $\sigma(t,x)$ be measurable, Lipschitz in $x$ uniformly in $t$, and of linear growth ($\sigma\equiv0$ is allowed).
Let the scalar process $X$ solve $\mathrm{d}X_t=b(t,X_t)\,\mathrm{d}t+\sigma(t,X_t)\,\mathrm{d}W_t$ on $[0,T]$, with $X_0$ square integrable and independent of $W$.
Then for all $0\le s\le t\le T$ and all non-decreasing $\varphi,\psi:\R\to\R$ with $\varphi(X_s),\psi(X_t)\in L^2$,
\begin{equation*}
\operatorname{Cov}\big(\varphi(X_s),\psi(X_t)\big)\ge0 .
\end{equation*}
\end{proposition}

\begin{corollary}
\label{cor:pos}
(i) With $\varphi=\psi=\mathrm{id}$, $\operatorname{Cov}(X_s,X_t)\ge0$; hence, on any grid, every such model has $\epsC\ge\|C^{\star}_{-}\|_F/\|C^{\star}\|_F$, where $C^{\star}_{-}=\min(C^{\star},0)$ entrywise.
(ii) If the marginal distribution functions $F_s,F_t$ are continuous, then with $\varphi=F_s$ and $\psi=F_t$ the lagged Spearman rank correlation $\rho_S(s,t)=12\operatorname{Cov}(F_s(X_s),F_t(X_t))$ is non-negative, and the bound also holds for $\rho_S$.
\end{corollary}

The result follows from the comparison theorem for scalar SDEs \citep{ikeda1977comparison}, the Markov property and Chebyshev's association inequality \citep{esary1967association}; see also \citet{herbst1991diffusion}.
In one dimension, two paths driven by the same noise cannot overtake each other without meeting, and once they meet they move together.
A particle that starts higher therefore stays at least as high, and Chebyshev's inequality turns this ordering into a non-negative covariance.
A negative lagged covariance needs the opposite: a particle that is high now and low later, as in an oscillator swinging through its equilibrium.
This requires information that $x$ does not carry, such as a velocity.
The hidden coordinates of the lift--project SDE carry it, so its $x$-paths can cross.
Negative lagged covariance in the data thus shows that $x$ alone is not a Markovian state, and it sets an error floor for every observed-state model.
The inverse problem is ill-posed: by Lemma~\ref{prop:nonid} the marginals never determine the coupling, and hidden coordinates are determined only up to a change of variables.
As in regularized inverse problems, the modeling prior selects a solution: the observed-state prior excludes every coupling with negative lagged covariance, whereas the lift--project prior selects that of a low-dimensional autonomous system, which is correct when the data come from one; experiments with common dynamics narrow the choice further.
The result covers the continuous-time versions of all our baselines: obs-TH, obs-TD, SB, and the deterministic OT-CFM flow ($\sigma\equiv0$).
An Euler--Maruyama step need not preserve the order of particles, so their discretizations obey it only approximately.
For a linear oscillator observed only in position, where every lagged covariance is exact, the diffusion that is Markovian in $x$ that mimics the marginals has exactly the rank-preserving coupling, whereas a lift--project model with one hidden coordinate reproduces the true coupling.

\paragraph{Several experiments with common dynamics.}
Suppose several experiments share the same dynamics but have different initial laws \(\mu_j\). Experiment \(j\) provides the observed marginal \(p_t^{(j)}\) obtained by evolving \(\mu_j\) under the common transition semigroup \(P_t\).
If each experiment is fitted with its own model, the experiments impose no constraints on one another. If one model must explain all experiments, it must fit every experiment at once, which can rule out models that fit each experiment alone.
In particular, consider two experiments with the same initial law for \(x\) but different later-time marginals. No model that is Markovian in \(x\) alone can fit both, because the same initial law and transition law must produce the same future marginal. A single model can fit both only by carrying additional information, such as a hidden initial state.

\section{Experiments}
\label{sec:exp}

\begin{figure}[t]
\centering
\includegraphics[width=0.76\textwidth]{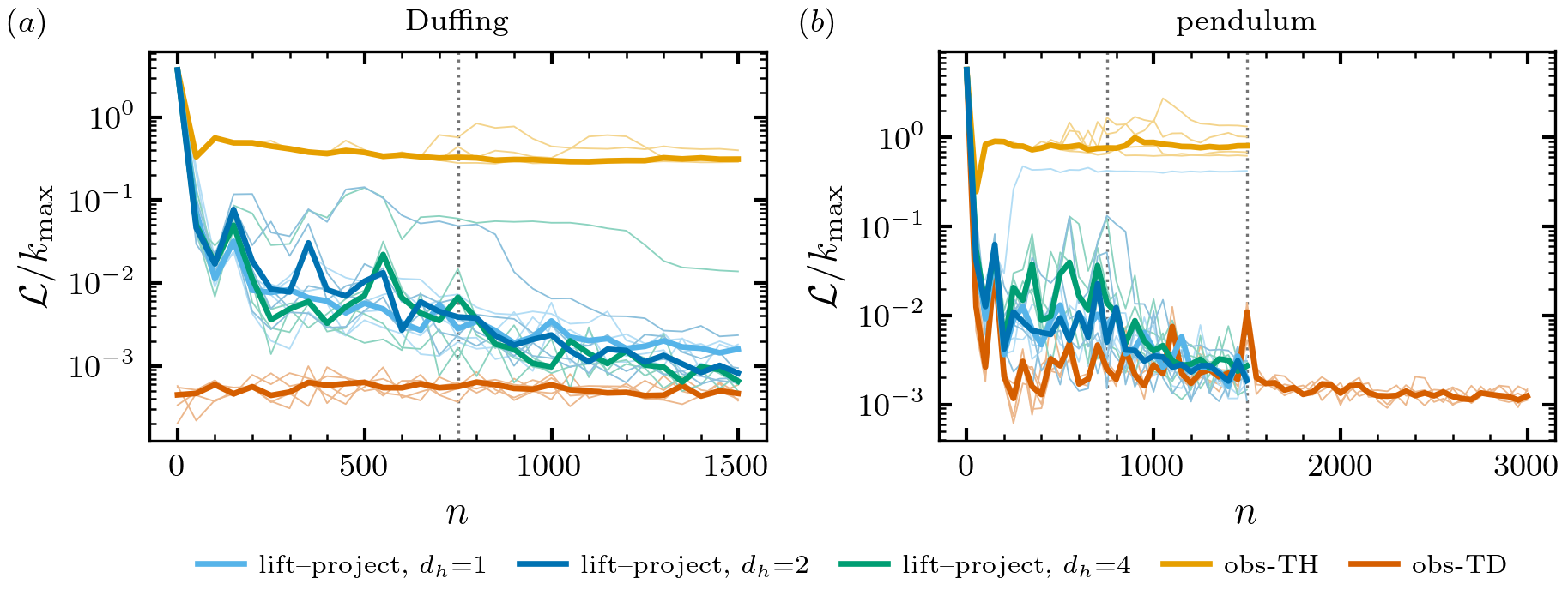}
\caption{Training curves on (a) Duffing and (b) the pendulum: minibatch loss per snapshot time against iteration $n$; thin lines are seeds and bold lines the seed median.
Dotted lines mark where the rollout horizon reaches $T$.
The lift--project loss falls by orders of magnitude for every hidden dimension, whereas obs-TH, which has no hidden state, stays far above it; the Duffing obs-TD curve starts low because it continues from a multiple-shooting solution.}
\label{fig:train}
\end{figure}

\paragraph{Systems and data.}
Our first example is the stochastic Duffing oscillator $\mathrm{d}x=v\,\mathrm{d}t$, $\mathrm{d}v=(-\gamma v-\alpha x-\beta x^3)\,\mathrm{d}t+\sigma\,\mathrm{d}W$ with $\gamma=0.2$, $\alpha=1$, $\beta=0.5$, $\sigma=0.15$ and $x_0\sim\mathcal N(1.5,0.2^2)$, $v_0\sim\mathcal N(0,0.2^2)$, and only the position $x$ is observed.
The second example is the damped pendulum $\mathrm{d}x=\omega\,\mathrm{d}t$, $\mathrm{d}\omega=(-0.1\,\omega-2\sin x)\,\mathrm{d}t+0.15\,\mathrm{d}W$, started near $103^\circ$ where it is strongly nonlinear, and only the angle $x$ is observed.
In both systems, the velocity is hidden, so the true hidden dimension is one.
Each training set has $16$ snapshots at $t=0,1,\dots,15$ ($T=15$) of $2000$ fresh particles, and evaluation uses $4000$ fresh trajectories.
We train five seeds per class and report seed medians; figures show all seeds.
Figure~\ref{fig:train} shows the training curves.

\subsection{Lagged covariance at comparable marginal fit}
\label{sec:select}

\begin{figure}[!t]
\centering
\includegraphics[width=0.94\textwidth]{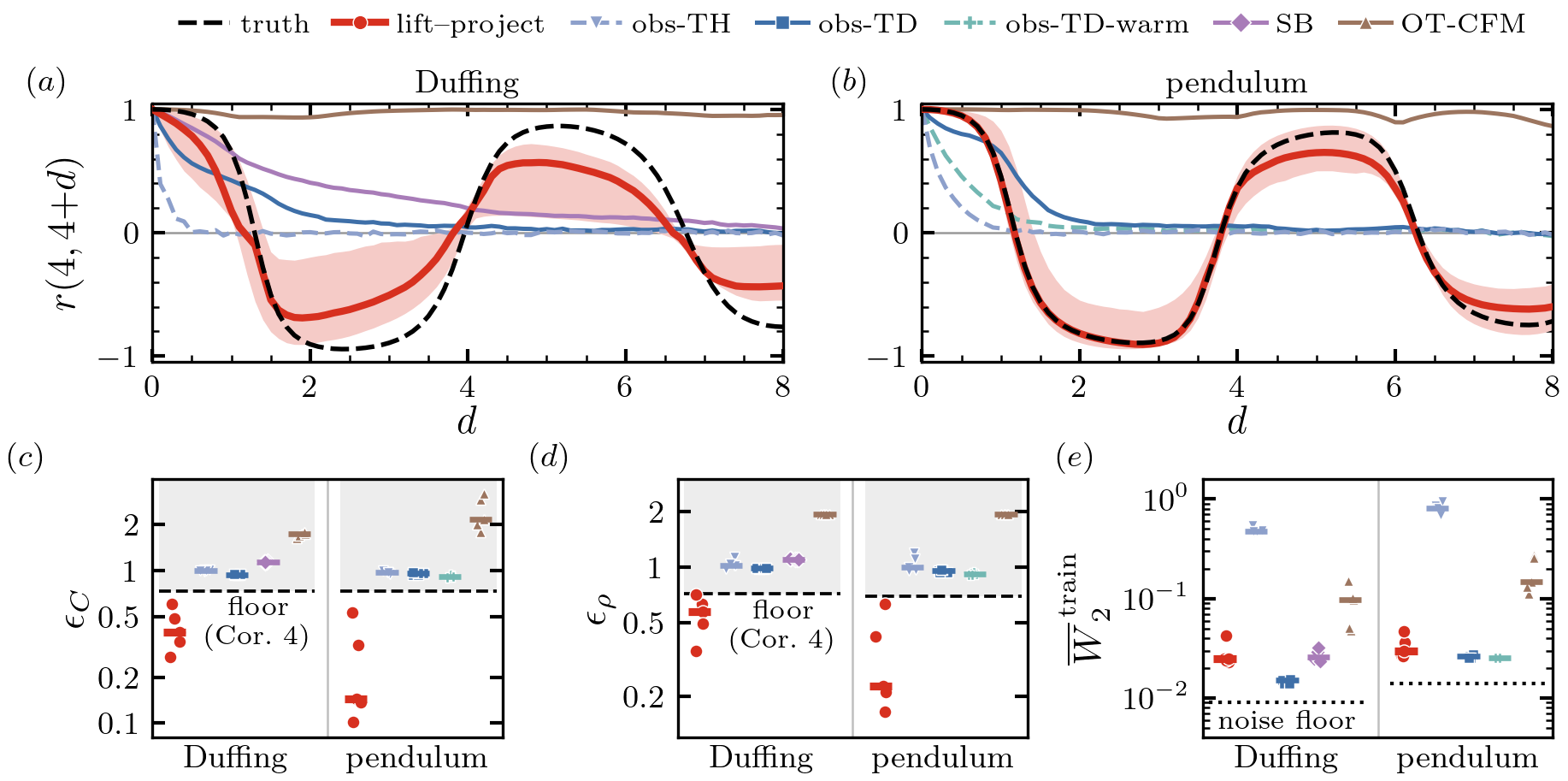}
\caption{At comparable marginal fit, the lift--project SDE is the only model below the error floor of observed-state models.
(a, b) Lagged correlation $r(4,4+d)=C(4,4+d)/\sqrt{C(4,4)\,C(4{+}d,4{+}d)}$: truth (dashed), lift--project seed median with its seed range shaded, and the seed median of each baseline.
(c) Relative lagged-covariance error $\epsC$. (d) Relative lagged-rank-correlation error $\epsilon_\rho$. (e) Training-window marginal error $\overline W_2^{\,\mathrm{train}}$.
Points: seeds; bars: seed medians; dashed lines: error floor for every diffusion that is Markovian in $x$ (grey: the region such models can reach); dotted: noise floors.}
\label{fig:adv}
\end{figure}

In both systems the true lagged correlation becomes strongly negative.
The lift--project SDE follows these negative lobes, whereas the observed-state models decay toward zero or stay near one (Figure~\ref{fig:adv}a,b).
Its lagged-covariance error is $2.4$ times lower than that of the strongest baseline on Duffing and $6.6$ times lower on the pendulum.
Every observed-state model stays above the error floor of Corollary~\ref{cor:pos}, and only the lift--project SDE falls below it, for both the lagged covariance and the lagged rank correlation (Figure~\ref{fig:adv}c,d).
The gap does not come from a better fit to the snapshots: obs-TD and SB fit them as well as the lift--project SDE does (Figure~\ref{fig:adv}e).
On the pendulum, the lift--project SDE also extrapolates better beyond the training window, with about a third of the marginal error of the best baseline.
Retraining on process-B data, or on another process with the same marginals, returns the same models: the lift--project SDE stays near the true process, and OT-CFM lands on process B (Appendix~\ref{app:processC}).

\paragraph{One hidden dimension is enough.}
\label{sec:dh}

\begin{figure}[!htb]
\centering
\includegraphics[width=0.78\textwidth]{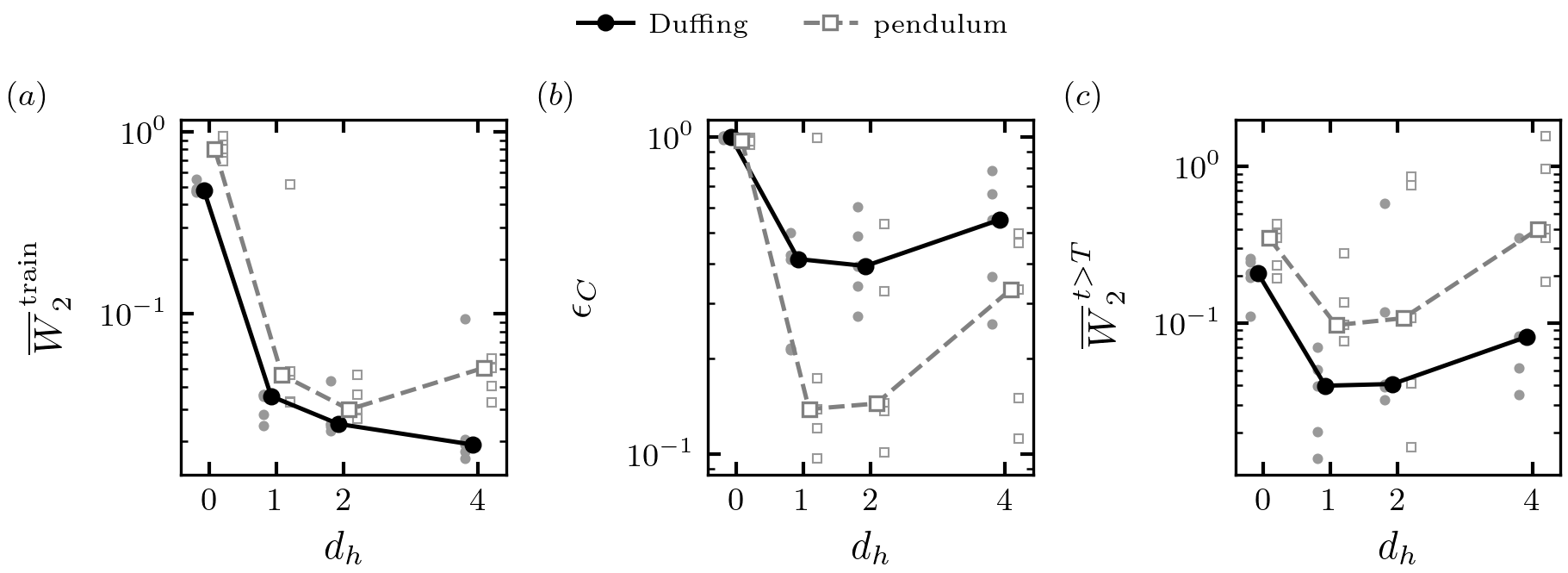}
\caption{Effect of the hidden dimension $d_h$ ($d_h=0$ is obs-TH).
(a) Training-window marginal fit.
(b) Relative lagged-covariance error $\epsC$ inside the observation window.
(c) Mean $W_2$ at $t=1.5T,2T,3T$ beyond the training window.
Large markers joined by lines are seed medians; small markers are the $5$ seeds.}
\label{fig:dh}
\end{figure}

The memory embedding suggests that the number of hidden coordinates matters more than the network size, so we varied $d_h\in\{0,1,2,4\}$ at a fixed parameter budget (Figure~\ref{fig:dh}).
One hidden coordinate, the true number, gives most of the gain in both the lagged-covariance error and the marginal fit, and more hidden coordinates do not help further.

\subsection{A second experiment favors hidden-state dynamics}
\label{sec:multi}

\begin{figure}[!t]
\centering
\includegraphics[width=0.84\textwidth]{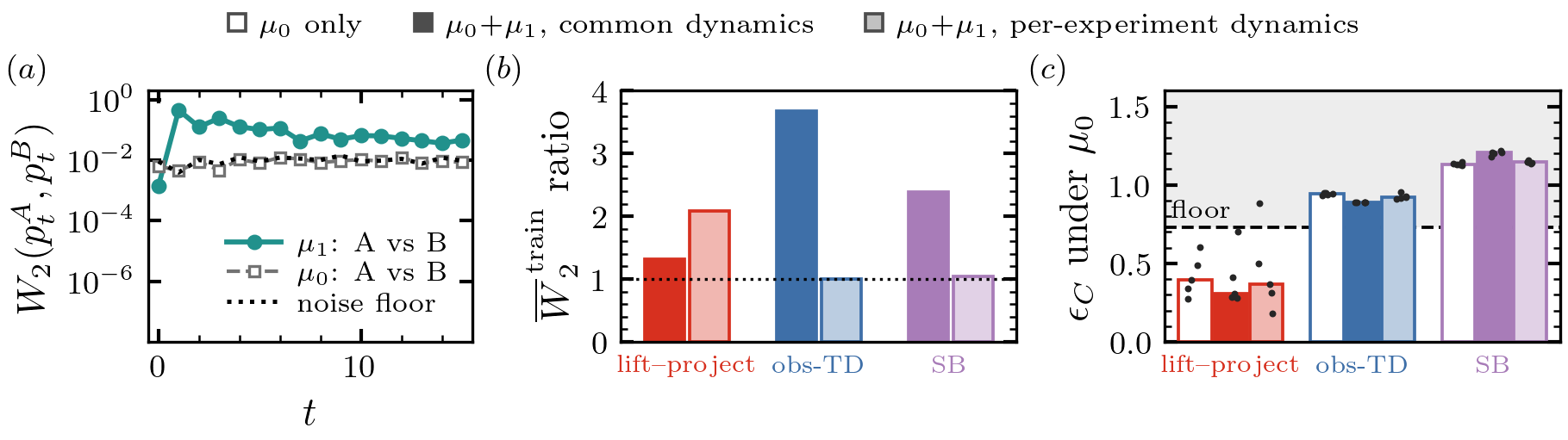}
\vspace{-5mm}
\caption{Two Duffing experiments with the same law of $x_0$ but different hidden velocities.
(a) $W_2$ between the marginals of the true process (A) and process B under $\mu_0$ and $\mu_1$, with the noise floor.
(b) Training marginal error with both experiments over that with one (worse experiment shown); dotted: equal fit.
(c) $\epsC$ under $\mu_0$; bars: seed medians, dots: seeds; dashed: error floor of observed-state models.}
\label{fig:multi}
\end{figure}

We added a second Duffing experiment $\mu_1$ with the same law of $x_0$ as $\mu_0$ but a shifted hidden velocity, $v_0\sim\mathcal N(1,0.2^2)$ instead of $\mathcal N(0,0.2^2)$.
The true process and process B have the same marginals under $\mu_0$.
Under $\mu_1$, with process B extended as the only version that is Markovian in $x$ (Appendix~\ref{app:details}), their marginals differ far beyond the noise floor (Figure~\ref{fig:multi}a).

We then trained the lift--project SDE, obs-TD and SB on both experiments in two ways (Figure~\ref{fig:multi}b).
With \emph{common dynamics}, one drift and diffusion must explain both experiments; for the lift--project SDE the experiment label enters only the initial hidden law.
With \emph{per-experiment dynamics}, the experiment label is also an input to the drift and diffusion, so each experiment can have its own dynamics.
As the theory predicts, observed-state models with common dynamics could not fit both experiments: their marginal error was several times that of single-experiment runs (Figure~\ref{fig:multi}b).
The lift--project SDE with common dynamics fitted both, and its lagged-covariance error stayed within the seed spread (Figure~\ref{fig:multi}c).
Models with per-experiment dynamics kept the coupling of their prior; the lift--project SDE among them fitted less well, suggesting it needs a longer training budget.

\section{Discussion}
\label{sec:discussion}

Snapshots leave the temporal coupling open, and the modeling prior fills it.
For partially observed physical systems, the lift--project SDE supplies the prior that realization theory suggests, a few hidden coordinates evolving autonomously, and it reduced the lagged-covariance error several-fold at comparable marginal fit.
Because marginal fit cannot tell couplings apart, it should be reported together with a two-time statistic, such as a lagged rank correlation.

Natural next steps are vector observations, a data-driven choice of the hidden dimension, and applications to single-cell and particle-tracking data.

\paragraph{Limitations.}
The coupling the lift--project SDE returns is set by its prior: it stayed near the true process when trained on marginal-equivalent data, so it would not recover a true system whose coupling is far from that of a low-dimensional autonomous system, such as process B.
It did not clearly improve the response to shifting every particle at $t=T/2$ (Appendix~\ref{app:details}).
On two FUCCI single-cell data sets, entropic OT predicted the withheld cell-cycle phase better, mainly because its paths pass through the training cells; on unseen cells its advantage is small (Appendix~\ref{app:fucci}).
Both synthetic systems have one hidden velocity; other hidden variables remain untested.

\clearpage
\acks{This work was supported by the National Science Foundation under award 2431610 and by the Air Force Office of Scientific Research under award FA9550-23-1-0405.
The authors used AI assistants (Claude, Anthropic; ChatGPT, OpenAI) to help write, debug, and comment code, improve figure quality, and modify the writing style for the machine learning community.
The authors reviewed all AI-assisted output and take full responsibility for the content of this paper.
Code to reproduce the results is available at \url{https://github.com/Oliverpoc/lift-project-sde}.}

\bibliography{references}

\clearpage
\appendix
\setcounter{topnumber}{3}\setcounter{totalnumber}{4}\renewcommand{\topfraction}{0.9}\renewcommand{\textfraction}{0.1}\renewcommand{\floatpagefraction}{0.7}

\section{Proofs}
\label{app:proofs}

\paragraph{Proof of Lemma~\ref{prop:nonid}.}
(i) Since $F_0$ is continuous, $U=F_0(X_0)$ is uniform on $(0,1)$.
With $Q_t(u)=\inf\{x:F_t(x)\ge u\}$ we have $Q_t(u)\le x$ if and only if $u\le F_t(x)$, so $\mathbb P(Y_t\le x)=\mathbb P(U\le F_t(x))=F_t(x)$.

(ii) For two square-integrable random variables $X,X'$ with joint distribution function $H$ and marginals $F,G$, Hoeffding's identity reads
\begin{equation*}
\operatorname{Cov}(X,X')=\int_{\R^2}\big(H(x,y)-F(x)G(y)\big)\,\mathrm{d}x\,\mathrm{d}y .
\end{equation*}
To see it, let $(\tilde X,\tilde X')$ be an independent copy of $(X,X')$, so that $2\operatorname{Cov}(X,X')=\E[(X-\tilde X)(X'-\tilde X')]$.
Write $X-\tilde X=\int_{\R}(\mathbf 1\{\tilde X\le x\}-\mathbf 1\{X\le x\})\,\mathrm{d}x$ and $X'-\tilde X'=\int_{\R}(\mathbf 1\{\tilde X'\le y\}-\mathbf 1\{X'\le y\})\,\mathrm{d}y$; the product of the two integrands has expectation $2(H(x,y)-F(x)G(y))$, and Fubini's theorem gives the identity.
Every joint distribution function satisfies $H(x,y)\le\min(F(x),G(y))$.
The pair $(Y_s,Y_t)=(Q_s(U),Q_t(U))$ attains this bound, because by the equivalence in (i), $\mathbb P(Y_s\le x,\,Y_t\le y)=\mathbb P(U\le F_s(x),\,U\le F_t(y))=\min(F_s(x),F_t(y))$.
Since $(X_s,X_t)$ and $(Y_s,Y_t)$ have the same marginals, the identity gives $\operatorname{Cov}(Y_s,Y_t)\ge\operatorname{Cov}(X_s,X_t)$.

(iii) Equality holds if and only if the joint distribution function of $(X_s,X_t)$ equals $\min(F_s(x),F_t(y))$ for almost every $(x,y)$, and hence, by right-continuity, for every $(x,y)$; that is, if and only if $(X_s,X_t)$ has the law of $(Q_s(U),Q_t(U))$.

\paragraph{Proof of Lemma~\ref{prop:gle}.}
The system $(x,v,u_1,\dots,u_m)$ has drift and diffusion that do not depend on $t$ and are Lipschitz, so it is an autonomous Markovian SDE with a unique strong solution.
Equation~\eqref{eq:aux} is linear in $u_i$, so
\begin{equation*}
u_i(t)=e^{-\lambda_i t}u_i(0)-c_i\int_0^t e^{-\lambda_i(t-s)}v(s)\,\mathrm{d}s+\varsigma\sqrt{2\lambda_i}\int_0^t e^{-\lambda_i(t-s)}\,\mathrm{d}W_i(s).
\end{equation*}
Inserting this into $\dot v=-\nabla U(x)+\sum_i c_iu_i$ gives $\dot v=-\nabla U(x)-\int_0^t\mathcal K(t-s)v(s)\,\mathrm{d}s+\tilde\xi(t)$, with
\begin{equation*}
\tilde\xi(t)=\sum_{i=1}^m c_i\,\eta_i(t),\qquad \eta_i(t)=e^{-\lambda_i t}u_i(0)+\varsigma\sqrt{2\lambda_i}\int_0^t e^{-\lambda_i(t-s)}\,\mathrm{d}W_i(s).
\end{equation*}
Each $\eta_i$ is a centered Gaussian process, and the $\eta_i$ are independent of each other and of $(x_0,v_0)$.
For $s\le t$, by the It\^o isometry,
\begin{equation*}
\E[\eta_i(t)\eta_i(s)]=\varsigma^2e^{-\lambda_i(t+s)}+2\lambda_i\varsigma^2\int_0^s e^{-\lambda_i(t-r)}e^{-\lambda_i(s-r)}\,\mathrm{d}r=\varsigma^2e^{-\lambda_i(t-s)} .
\end{equation*}
Hence $\E[\tilde\xi(t)\tilde\xi(s)]=\varsigma^2\sum_ic_i^2e^{-\lambda_i|t-s|}=\varsigma^2\mathcal K(|t-s|)$, so $\tilde\xi$ and $\xi$ are centered Gaussian processes with the same covariance, and $(x_0,v_0,\tilde\xi)$ has the same law as $(x_0,v_0,\xi)$.
For a given initial state and forcing path, \eqref{eq:gle} is a Lipschitz integro-differential equation with a unique solution, which is a measurable function of the initial state and the forcing.
The $x$-component of the Markovian system and the solution of \eqref{eq:gle} are therefore the same function of two inputs with the same law, and so have the same law.

The construction uses two assumptions: the $u_i$ start from their stationary law, and the noise covariance is tied to the memory kernel, $\E[\xi(t)\xi(s)]=\varsigma^2\mathcal K(|t-s|)$.
Without the second, the same auxiliary variables still give a finite-dimensional Markovian model, with a different noise term.
A kernel that is not a finite sum of exponentials can be approximated by one.

\paragraph{Proof of Proposition~\ref{prop:pos}.}
Under the Lipschitz and linear-growth conditions the SDE has a unique strong solution with $\E\sup_{t\le T}X_t^2<\infty$ \citep[Ch.~5]{karatzas1991brownian}.
Fix $0\le s\le t\le T$ and let $X^{x}$ be the solution on $[s,T]$ started at $X^{x}_s=x$ and driven by the same Brownian motion.
It is a measurable function of $x$ and of the increments $(W_r-W_s)_{r\ge s}$, which are independent of $X_s$, and by uniqueness $X_t=X^{X_s}_t$.

\emph{Step 1: paths do not cross.}
Let $x<y$ and let $\tau$ be the first time $r\ge s$ with $X^x_r=X^y_r$ ($\tau=T$ if there is none).
By continuity, $X^x_r<X^y_r$ for $r<\tau$.
The path that equals $X^y$ on $[s,\tau]$ and $X^x$ on $[\tau,T]$ also solves the SDE from $y$, because $\tau$ is a stopping time; by uniqueness it equals $X^y$.
Hence $X^x_t\le X^y_t$ almost surely.
This is the comparison theorem of \citet{ikeda1977comparison} for equal coefficients.

\emph{Step 2: the conditional expectation is non-decreasing.}
Write $\psi=\psi^+-\psi^-$ with $\psi^+=\max(\psi,0)$ non-decreasing and $\psi^-=\max(-\psi,0)$ non-increasing, and let $g^\pm(x)=\E\,\psi^\pm(X^x_t)\in[0,\infty]$.
By Step~1, $g^+$ is non-decreasing and $g^-$ is non-increasing.
Since the increments after $s$ are independent of $X_s$, $\E[\psi^\pm(X_t)\mid X_s]=g^\pm(X_s)$, and since $\E\,g^\pm(X_s)=\E\,\psi^\pm(X_t)<\infty$, both are finite on a set $S$ with $\mathbb P(X_s\in S)=1$.
On $S$ the function $g=g^+-g^-$ is non-decreasing, $\E[\psi(X_t)\mid X_s]=g(X_s)$, and $\E\,g(X_s)^2\le\E\,\psi(X_t)^2<\infty$ by Jensen's inequality.

\emph{Step 3: association.}
By the tower property, $\operatorname{Cov}(\varphi(X_s),\psi(X_t))=\operatorname{Cov}(\varphi(X_s),g(X_s))$.
With $X'_s$ an independent copy of $X_s$,
\begin{equation*}
2\operatorname{Cov}\big(\varphi(X_s),g(X_s)\big)=\E\big[(\varphi(X_s)-\varphi(X'_s))\,(g(X_s)-g(X'_s))\big]\ge0,
\end{equation*}
because $\varphi$ and $g$ are both non-decreasing, so the two factors never have opposite signs.
This is Chebyshev's association inequality \citep{esary1967association}.
For $s=t$, Step~3 applies directly with $g=\psi$.

\paragraph{Proof of Corollary~\ref{cor:pos}.}
(i) Take $\varphi=\psi=\mathrm{id}$; $X_s,X_t\in L^2$ by the moment bound above.
If every entry of $\hat C$ is non-negative, then $|\hat C_{ij}-C^\star_{ij}|\ge|\min(C^\star_{ij},0)|$ for every grid entry, and summing squares gives $\|\hat C-C^\star\|_F\ge\|C^\star_-\|_F$.

(ii) $F_s$ and $F_t$ are bounded and non-decreasing.
If they are continuous, $F_s(X_s)$ and $F_t(X_t)$ are uniform on $(0,1)$ with variance $1/12$, so $\rho_S(s,t)=12\operatorname{Cov}(F_s(X_s),F_t(X_t))\ge0$, and the bound follows as in (i).

The bound is the distance from $C^\star$ to the nearest matrix with no negative entries, and a diffusion need not attain it.
It holds for continuous-time models; an Euler--Maruyama step $x\mapsto x+b(t,x)\Delta t+\sigma(t,x)\Delta W$ need not be non-decreasing in $x$, so Step~1 can fail after discretization (Appendix~\ref{app:details} gives an example).

\section{Experimental details}
\label{app:details}

\paragraph{Process B in the experiments.}
In the experiments, $F_0$ and $Q_t$ are the empirical distribution and quantile functions of a $40{,}000$-particle sample of the true process (Duffing) or of its evaluation sample (pendulum); process-B samples are $x_t=Q_t(F_0(x_0))$ with new draws of $x_0$.

\paragraph{Extension of process B to a second experiment.}
Process B is defined for $\mu_0$.
Under $\mu_1$ we use its only version that is Markovian in $x$: each particle moves deterministically by $x_s\mapsto Q_t(F_s(x_s))$, which is well defined because $\mu_1$ and $\mu_0$ have the same law of $x_0$.
Figure~\ref{fig:multi}(a) therefore separates the true process from this version; versions of B with hidden state also exist.

\paragraph{Training.}
Algorithm~\ref{alg:train} gives the training loop of the lift--project SDE; obs-TH is the same loop with $d_h=0$ and no generator.
All SDE models use Euler--Maruyama with step $\Delta t=0.1$.
The drift and diffusion network of the lift--project SDE is an MLP with two hidden layers of width $64$ and SiLU activations; it maps $z\in\R^{1+d_h}$ to a drift and a diagonal diffusion, and its last layer is initialized at $0.1$ times the default scale.
On the pendulum the width is $65$, $64$ and $60$ for $d_h=1,2,4$; obs-TH has width $76$ (Duffing) or $75$ (pendulum), so that parameter counts match.
The generator $G_\phi$ is an MLP with two hidden layers of width $32$ and SiLU activations; its input is $(x_0,\varepsilon)$ with $\varepsilon\sim\mathcal N(0,I_{d_h})$, and its output is $h_0$.
With two experiments, a one-hot experiment label is an extra input of the generator (common dynamics) or of the generator, drift and diffusion (per-experiment dynamics).
The lift--project SDE and obs-TH are trained for $1500$ Adam iterations with $1000$ particles per minibatch, a learning rate decayed from $3\times10^{-3}$ to $3\times10^{-4}$ by a cosine schedule, and gradient clipping at global norm $1$.
The rollout horizon grows linearly from $4$ to $T=15$ over the first half of the iterations and stays at $T$ afterwards; the loss uses every snapshot time up to the horizon.
Each run used one CPU process.

\paragraph{Baseline training.}
obs-TD uses Fourier time features $[t/T,\sin(j\pi t/T),\cos(j\pi t/T)]_{j\le 10}$ (Duffing) or $[t/T,\sin(2\pi jt/30),\cos(2\pi jt/30)]_{j\le10}$ (pendulum) and a width of $67$ (Duffing) or $66$ (pendulum).
On Duffing it first runs $3000$ multiple-shooting iterations, in which each interval $[t_k,t_{k+1}]$ is simulated from $200$ data samples at $t_k$ and compared with the snapshot at $t_{k+1}$, and then $1500$ full-rollout iterations with the curriculum and one third of the learning rate.
On the pendulum it runs $3000$ full-rollout iterations with the learning rate above and the curriculum over the first $1500$.
obs-TD-warm is first trained for $2000$ iterations to copy the OT-CFM drift of the same seed.
OT-CFM and SB use $20{,}000$ iterations with $67$ pairs per snapshot interval; SB uses reference noise $\sigma_b=0.25$.
For two experiments the batch is split into $500$ particles per experiment; with common dynamics obs-TD and SB receive no experiment label, and with per-experiment dynamics the label is an extra input of their networks.

\paragraph{Supplementary observations.}
obs-TH, which does not fit the marginals (Figure~\ref{fig:train}), has a few negative lagged-covariance entries on Duffing, because its drift is steep enough that one Euler--Maruyama step can swap the order of two particles; its rank-correlation error stays above the floor.
Across all seeds and training processes, SB and OT-CFM have no negative lagged-covariance entry on either system, and OT-CFM preserves ranks exactly ($\rho_S=1$).
obs-TD has a few entries just below zero, consistent with Proposition~\ref{prop:pos} up to discretization and sampling.
A regression of the hidden velocity on the learned $h$ fitted well even for untrained networks, so it cannot show whether $h$ learned the velocity.

\paragraph{Stability prior.}
Every learned drift $f$ is replaced by $B\tanh(f/B)-\kappa(|z|-R)_+\,z/|z|$ and the diffusion is bounded.
On Duffing $B=5$, $\kappa=1$, $R=4$ and the diffusion bound is $1$; on the pendulum $B=10$, $\kappa=5$, $R=8$ and the bound is $2$; $R$ is more than three times the data radius.
Outside radius $R+B\sqrt{n}/\kappa$, with $n=1+d_h$ the dimension of $z$, the drift points inward, so no path can escape to infinity.
In the pendulum case, one of the five $d_h=1$ seeds stalled; it is kept in all statistics and figures.

\paragraph{Metrics.}
$\overline W_2^{\,\mathrm{train}}$ is the mean $W_2$ over the $15$ training times after $t=0$; the held-out version uses the mid-times $t_k+0.5$; extrapolation uses $t=1.5T,2T,3T$; the intervention metric shifts every particle by $+0.6$ (Duffing) or $+0.4$ (pendulum) at $t=T/2$ and keeps $h$ unchanged.
Time-dependent baselines are extrapolated in two ways, by continuing the time input and by freezing it at $T$, and the better one is used.
Noise floors are computed from an independent truth sample.

\paragraph{Summary statistic.}
The main text reports seed medians, and every figure shows all five seeds; using seed means instead changes no comparison.

\section{The data sets}
\label{app:data}

\begin{figure}[htbp]
\centering
\includegraphics[width=\textwidth]{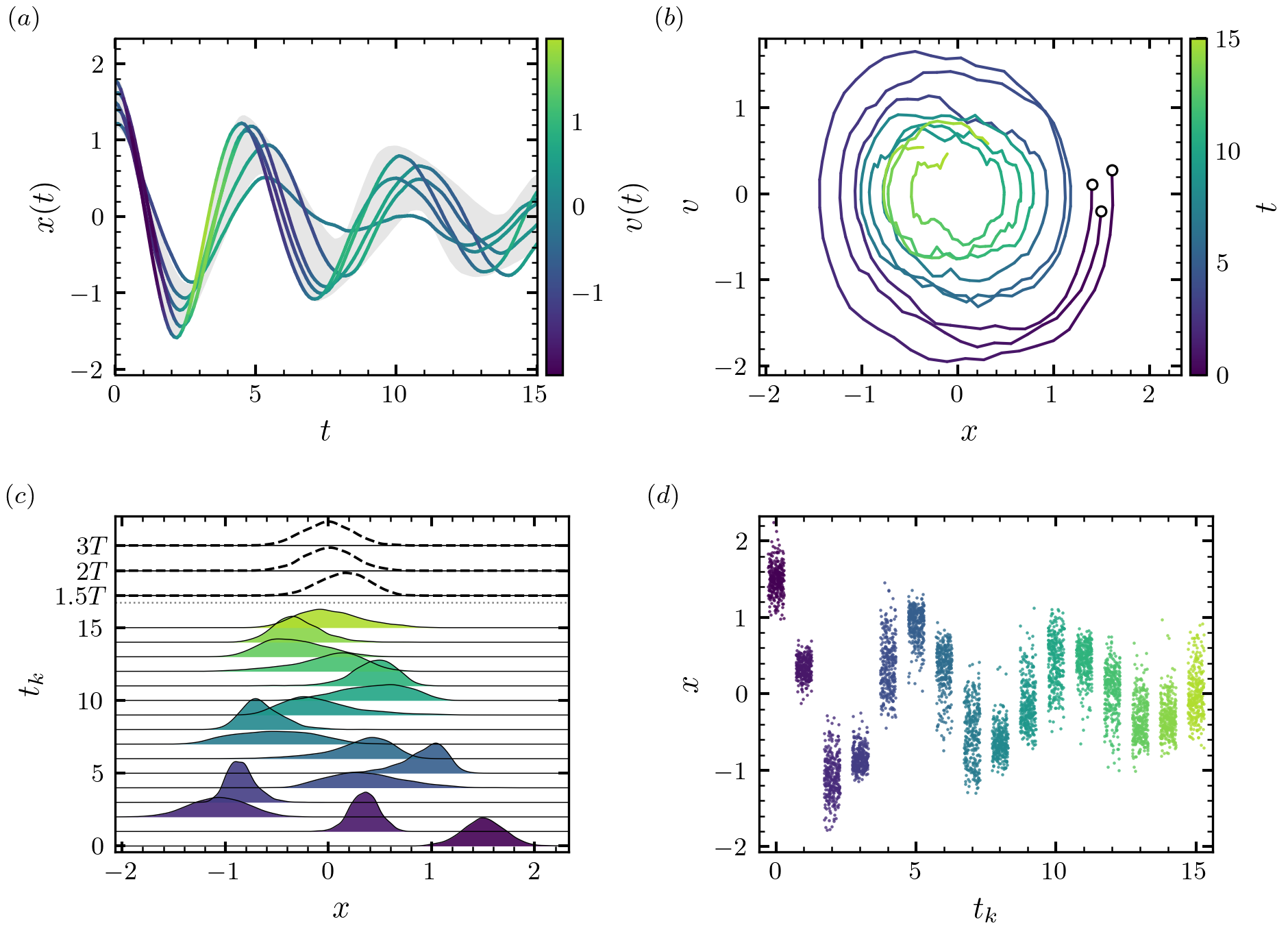}
\caption{Duffing data (the true process); only the position $x$ is observed.
(a) Five paths of the true system in $[0,T]$, colored by the hidden velocity $v$; grey: the $5$--$95\%$ range of the marginals.
(b) Phase portrait $(x,v)$ of three of these paths, colored by time $t$; circles mark $t=0$.
(c) Densities of $x$ at the $16$ training times (colored by $t_k$) and at the extrapolation times $1.5T$, $2T$, $3T$ (dashed, never used in training).
(d) The training data as a model sees them: $300$ of the $2000$ unpaired samples at each $t_k$, spread horizontally for visibility.}
\label{fig:data-duffing}
\end{figure}

\begin{figure}[htbp]
\centering
\includegraphics[width=\textwidth]{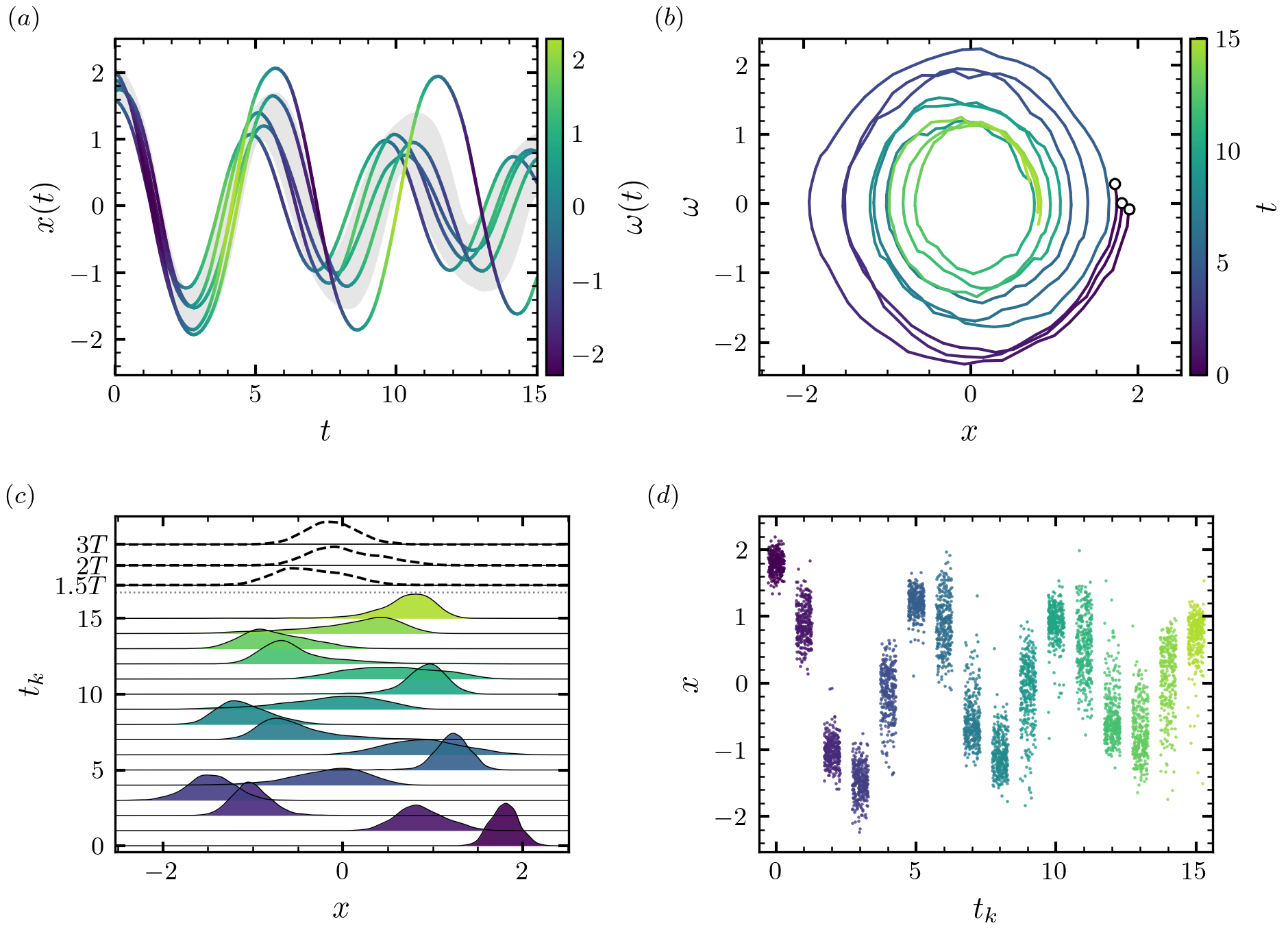}
\caption{Pendulum data (the true process); only the angle $x$ is observed.
(a) Five paths of the true system in $[0,T]$, colored by the hidden angular velocity $\omega$; grey: the $5$--$95\%$ range of the marginals.
(b) Phase portrait $(x,\omega)$ of three of these paths, colored by time $t$; circles mark $t=0$.
(c, d) As in Figure~\ref{fig:data-duffing}.}
\label{fig:data-pendulum}
\end{figure}

\begin{figure}[htbp]
\centering
\includegraphics[width=\textwidth]{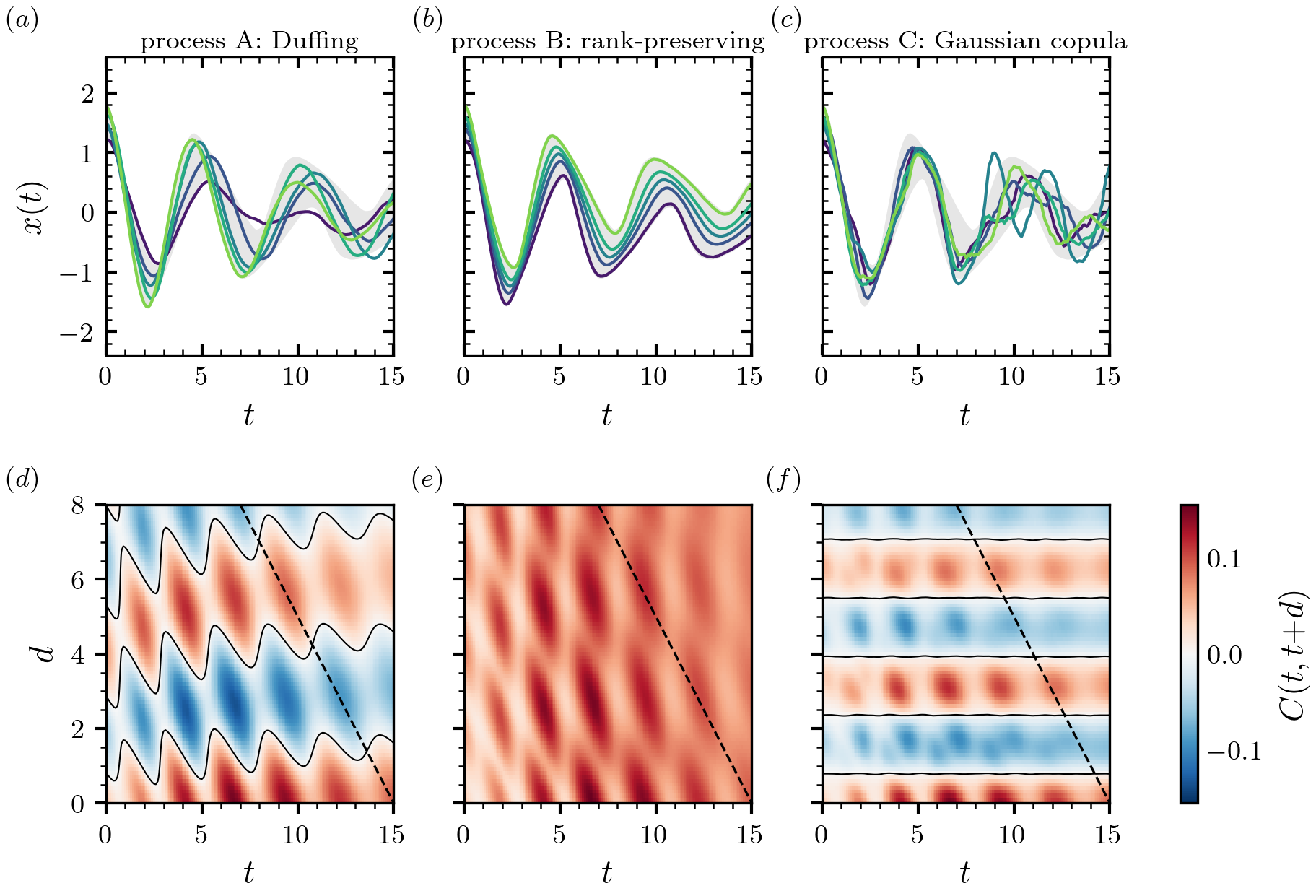}
\caption{Three Duffing processes with the same marginals.
(a--c) Five sample paths of the true process A (the stochastic Duffing oscillator), process B (the rank-preserving process of Lemma~\ref{prop:nonid}) and process C (the Gaussian-copula process of Appendix~\ref{app:processC}); the color of a path encodes the rank of its initial value $x_0$, and the grey band is the $5$--$95\%$ range of the shared marginals.
Process-B paths never cross, so their colors stay ordered; process-A and process-C paths cross.
(d--f) Lagged covariance $C(t,t+d)$ of each process from $4000$ paths, on one color scale centered at zero; solid line: $C=0$; dashed line: $t+d=T$, the edge of the $\epsC$ grid.}
\label{fig:data-processes}
\end{figure}

Figures~\ref{fig:data-duffing} and~\ref{fig:data-pendulum} show the two oscillators: panels (a) and (b) show the hidden velocity that makes paths cross, and panels (c) and (d) show what the models receive.
Figure~\ref{fig:data-processes} shows three Duffing processes with the same marginals, the true process (A), process B and process C: their paths fill the same marginal band, but only processes A and C have negative covariance.

\section{Training on marginal-equivalent data}
\label{app:processC}

\begin{figure}[t]
\centering
\includegraphics[width=\textwidth]{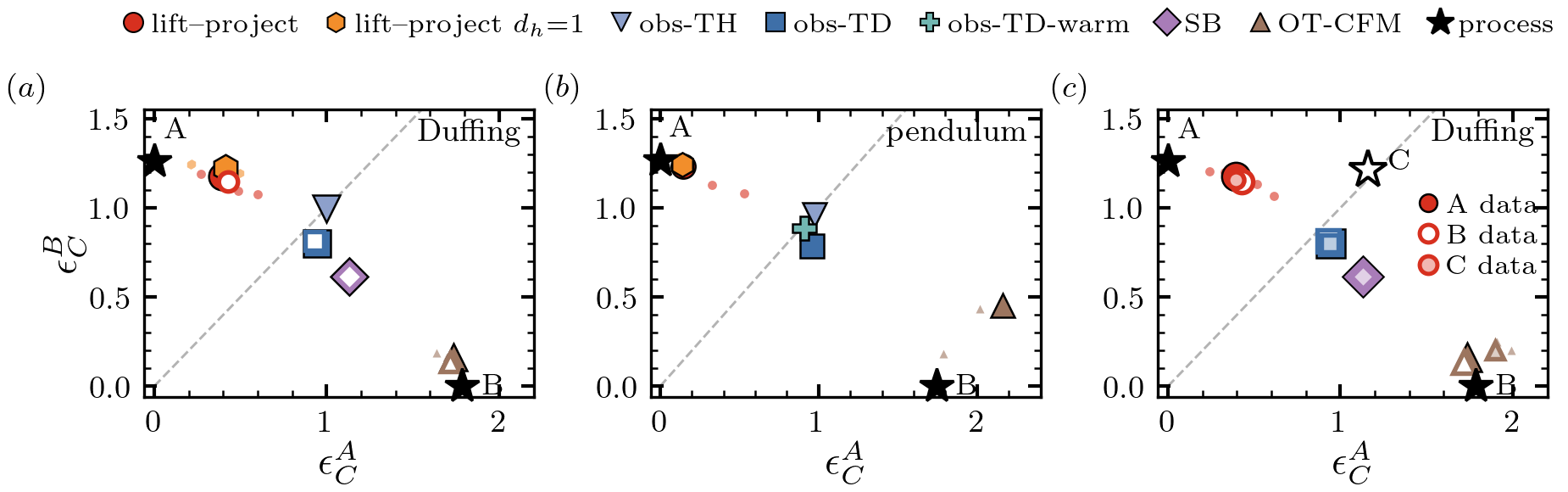}
\caption{Each model class selects a stable coupling across marginal-equivalent data.
Models are placed by their relative lagged-covariance error to process A ($\epsC^A$) and process B ($\epsC^B$); large markers: seed medians; small markers: individual seeds; stars: processes.
(a) Duffing, trained on process-A (filled) and process-B data (open).
(b) Pendulum, process-A data.
(c) Duffing, trained on A (filled), B (open) and C data (light); open star: process C, marginal-equivalent to A and B.}
\label{fig:select}
\end{figure}

To check where each modeling prior lands, we trained every class on the true process (here called process A), on process B, and on a third process C with the same marginals.
\emph{Process~C} lets the rank move, $x_t=Q_t(U_t)$ with $U_t$ uniform at every time; any such $U_t$ reproduces the marginals, so the lagged covariance can be made almost arbitrary.

The lift--project SDE lands near process A and OT-CFM near process B (Figure~\ref{fig:select}a,b).
Trained on process-B data, the lift--project SDE still lands near process A, and OT-CFM still lands on process B.
Trained on process-C data, every class stayed near its preferred coupling (Figure~\ref{fig:select}c): the lift--project SDE near A, the three baselines near B, and every seed closer to that process than to C.
Each class therefore returns the same coupling on all three data sets; for the lift--project SDE, it is the coupling of a low-dimensional autonomous system.

\paragraph{Construction of process C.}

Process C is $x_t=Q_t(\Phi(Y_t))$, where $Q_t$ is the quantile function of the true process and $\Phi$ the standard normal distribution function.
$Y$ is the first component of the two-dimensional linear SDE $\mathrm{d}\mathbf Y=(-I/\tau+\Omega J)\mathbf Y\,\mathrm{d}t+\sqrt{2/\tau}\,\mathrm{d}\mathbf W$ with $\mathbf Y_0\sim\mathcal N(0,I)$ and $J$ the $90^\circ$ rotation matrix; $Y$ alone is not Markovian, because its oscillating correlation needs the second component.
Then $Y_t\sim\mathcal N(0,1)$ at every $t$, $\Phi(Y_t)$ is uniform, and $\operatorname{corr}(Y_t,Y_{t+d})=e^{-d/\tau}\cos(\Omega d)$, so with exact quantile functions $x_t$ has exactly the marginals of the true process.
Figure~\ref{fig:data-processes} shows process-C paths and their lagged covariance next to those of processes A and B.
In the experiments $Q_t$ is the empirical quantile function of the $40{,}000$-particle sample of the true process that also defines process B; the largest marginal $W_2$ between processes C and A is then $0.011$, below the noise floor of $0.012$.
We chose $(\tau,\Omega)=(20,2)$ from a small grid before any training on process C, to put C far from A, from B and from where every class already lands, so that a model landing near C could not be explained by its prior.
The four classes were trained with the unchanged Duffing code and budgets.
Compared with A data, the medians barely moved, and the largest change, for OT-CFM, is comparable to its seed spread.

\begin{center}
\footnotesize\setlength{\tabcolsep}{4pt}
\begin{tabular}{lccc|ccc}
\toprule
& \multicolumn{3}{c|}{Median} & \multicolumn{3}{c}{Mean $\pm$ sd} \\
C data & $\epsC^A$ & $\epsC^B$ & $\epsC^C$ & $\epsC^A$ & $\epsC^B$ & $\epsC^C$ \\
\midrule
lift--project ($d_h=2$) & $0.392$ & $1.156$ & $1.190$ & $0.420\pm0.148$ & $1.153\pm0.057$ & $1.178\pm0.046$ \\
obs-TD & $0.938$ & $0.802$ & $1.148$ & $0.936\pm0.004$ & $0.805\pm0.004$ & $1.143\pm0.008$ \\
OT-CFM & $1.895$ & $0.203$ & $2.125$ & $1.854\pm0.114$ & $0.177\pm0.061$ & $2.073\pm0.139$ \\
SB & $1.131$ & $0.616$ & $1.387$ & $1.136\pm0.011$ & $0.613\pm0.008$ & $1.394\pm0.015$ \\
\bottomrule
\end{tabular}
\end{center}

\section{Real-data case study: different systems suit different priors}
\label{app:fucci}

Single-cell measurements are a natural source of snapshot data, and understanding the systems they describe requires linking data across scales \citep{chandrasekaran2021axes}.
Differential-equation and dynamic-network models are already used to infer developmental dynamics from time-course data \citep{schwartz2021yolk,schwartz2023dynamic}.
This case study asks whether the lift--project prior helps on cyclic single-cell data, and whether marginal fit alone can tell which prior suits the data.
An independent measure of cell-cycle position, the continuous phase of the FUCCI reporter \citep{sakaue2008visualizing}, is withheld from all models and used to judge the answer.
Entropic optimal transport predicted the withheld phase best.
On these data the transport prior suits the system better than the hidden-state prior.

\begin{figure}[htbp]
\centering
\includegraphics[width=\textwidth]{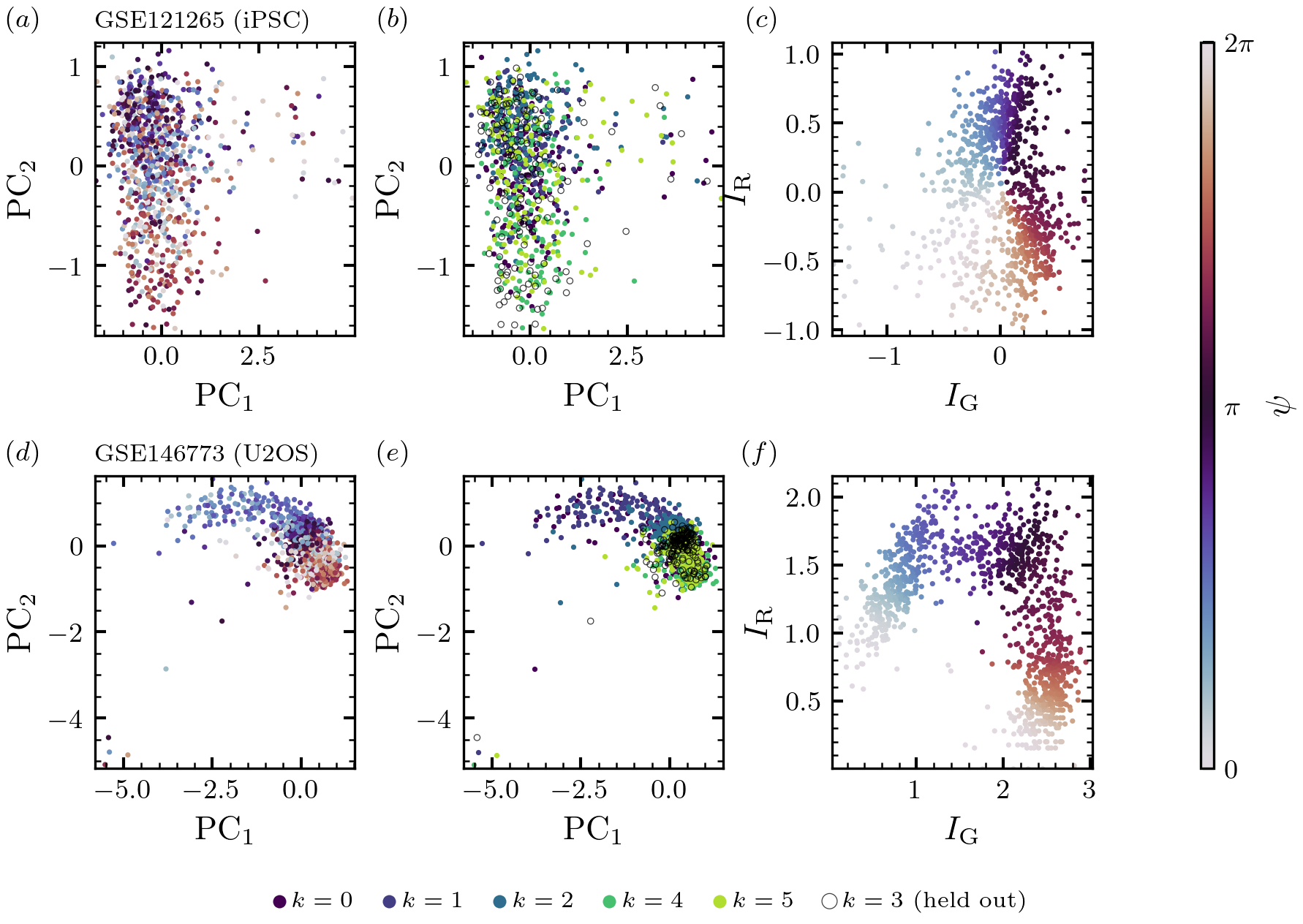}
\caption{The two FUCCI datasets; top row GSE121265 (iPSC, $888$ cells), bottom row GSE146773 (U2OS, $1146$ cells).
(a, d) The first two of the six principal components that form the observed state $x$, colored by the continuous FUCCI phase $\psi$ (cyclic color scale).
(b, e) The same cells colored by the six equal-count phase bins $k$ that define the snapshots; open circles mark the held-out bin $k=3$.
(c, f) FUCCI reporter intensities, green channel $I_{\mathrm G}$ against red channel $I_{\mathrm R}$, colored by $\psi$ (GSE121265: background-adjusted $\log_{10}$ sums; GSE146773: rescaled log intensities, as provided by the authors).
Axes show the central $99\%$ of cells.
The continuous phase $\psi$ is shown here for illustration only; no model was trained on it.}
\label{fig:data-fucci}
\end{figure}

\paragraph{Data.}
GSE121265 \citep{hsiao2020characterizing} contains $888$ induced pluripotent stem cells with FUCCI imaging and single-cell RNA sequencing (Fluidigm C1, UMI counts).
GSE146773 \citep{mahdessian2021spatiotemporal} contains $1146$ U2OS cells with FUCCI coordinates and SMART-seq2 counts.
We removed the FUCCI transgene transcripts and spike-ins, normalized each cell to $10^4$ counts, applied $\log(1+\cdot)$, kept genes detected in at least $10\%$ of cells, centered each gene within its technical batch (individual or plate), and kept the $1000$ most variable genes.
The observed state $x\in\R^6$ is the top six principal components; no FUCCI information enters $x$.
The truth is the authors' continuous FUCCI phase, rescaled by its ranks to a phase $\psi\in[0,2\pi)$ that is uniform over the cells.
Figure~\ref{fig:data-fucci} shows both datasets.

\paragraph{Task.}
The cells are split into six equal-count bins of $\psi$, and bin $k$ is treated as a snapshot at model time $t_k=k+\tfrac12$, with period $6$; the continuous phase is withheld.
Bin $3$ is never loaded during training and serves as a held-out marginal.
Models start from bin-$0$ cells and are matched to the five training bins over two cycles.
Each cell is assigned the average model time, taken on the circle, of its $20$ nearest generated points, and this phase is compared with $\psi$ by the mean absolute angular error (MAE).

\paragraph{Models.}
The classes are the lift--project SDE ($d_h=2$ and $d_h=1$), obs-TH, obs-TD and OT-CFM, with the same stability prior.
The observed state is six-dimensional, so $W_2$ is computed by exact minibatch optimal transport (batch $256$, $2000$ iterations); there are three seeds per class, and parameter counts are matched within $2\%$ (about $6900$).
obs-TD receives periodic time features with the true period, which favors it.
In place of SB we match the cells of consecutive training bins by exact or entropic optimal transport and join matched cells by straight lines, as in Waddington-OT \citep{schiebinger2019optimal}; these OT chains learn no dynamics.
Proposition~\ref{prop:pos} concerns scalar $x$ and does not apply here.
Results are seed means.

\begin{figure}[h]
\centering
\includegraphics[width=\textwidth]{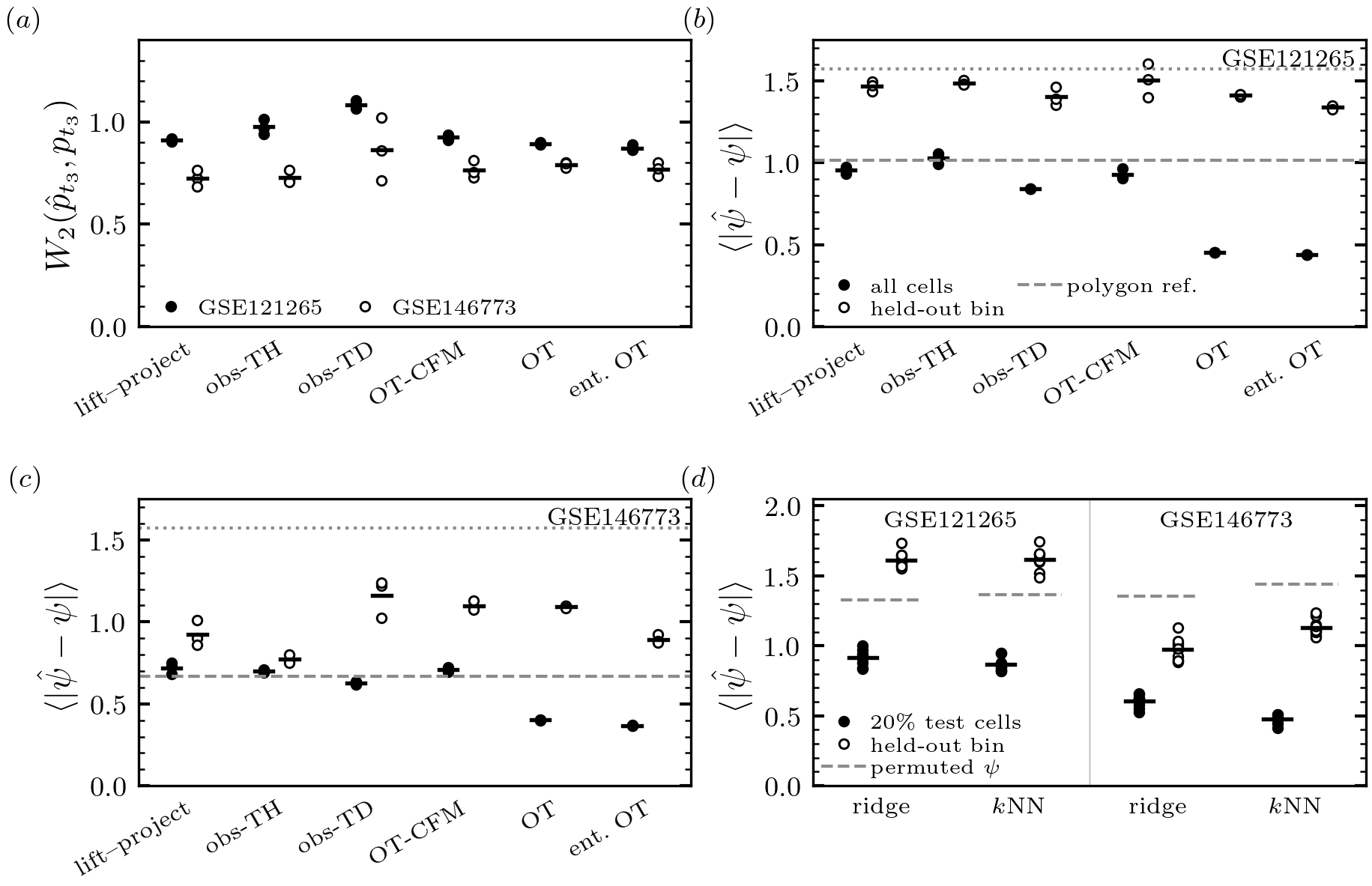}
\caption{FUCCI case study.
(a) Held-out-bin marginal fit $W_2$ for each class; filled: GSE121265, open: GSE146773.
(b, c) Circular mean absolute error against the withheld FUCCI phase $\psi$ for GSE121265 and GSE146773; filled: all cells, open: held-out-bin cells; dashed: centroid-polygon reference; dotted: chance ($\pi/2$).
(d) Supervised static readout from one transcriptome; filled: unseen cells of the training bins, open: held-out-bin cells; dashed: permuted labels.
Points are seeds (a--c) or random splits (d); bars are means.}
\label{fig:fucci}
\end{figure}

\paragraph{Results.}
The table and Figure~\ref{fig:fucci} give the outcome.
On the held-out bin, the lift--project SDE fitted the marginal about as well as the best class on GSE121265 and best on GSE146773.
On the training bins, the OT chains fit better, because their paths pass through the training cells themselves.
For the same reason entropic OT gave the smallest phase error on both datasets, about half that of the lift--project SDE (third limitation below).
A reference without dynamics, the centroid polygon, which projects each cell onto the closed polygon through the training-bin centers, was slightly worse than the lift--project SDE on GSE121265 and better on GSE146773.
On GSE146773 the autonomous models carried more paths through a full cycle than the OT chains and OT-CFM; since obs-TH did so too, this is not an effect of the hidden state.

\begin{center}
\footnotesize\setlength{\tabcolsep}{3.5pt}
\begin{tabular}{l|ccc|ccc}
\toprule
& \multicolumn{3}{c|}{GSE121265 (iPSC)} & \multicolumn{3}{c}{GSE146773 (U2OS)} \\
Method & held-out $W_2$ & MAE & held-out-bin MAE & held-out $W_2$ & MAE & held-out-bin MAE \\
\midrule
lift--project ($d_h=2$) & $0.911$ & $0.957$ & $1.468$ & $0.724$ & $0.719$ & $0.923$ \\
lift--project ($d_h=1$) & $0.959$ & $1.016$ & $1.460$ & $0.710$ & $0.740$ & $0.896$ \\
obs-TH & $0.976$ & $1.029$ & $1.487$ & $0.729$ & $0.701$ & $0.772$ \\
obs-TD & $1.084$ & $0.842$ & $1.402$ & $0.864$ & $0.628$ & $1.160$ \\
OT-CFM & $0.926$ & $0.929$ & $1.504$ & $0.765$ & $0.710$ & $1.096$ \\
exact OT chain & $0.894$ & $0.452$ & $1.413$ & $0.792$ & $0.402$ & $1.092$ \\
entropic OT chain & $0.872$ & $\mathbf{0.441}$ & $1.339$ & $0.770$ & $\mathbf{0.368}$ & $0.894$ \\
centroid polygon (reference) & -- & $1.013$ & $1.194$ & -- & $0.666$ & $0.845$ \\
\bottomrule
\end{tabular}
\end{center}

\paragraph{Static readout.}
To see how much phase the instantaneous transcriptome carries, we fitted a supervised map from $x$ to $(\cos\psi,\sin\psi)$ on $80\%$ of the training-bin cells, by ridge regression and by $k$-nearest-neighbour regression with hyperparameters chosen by cross-validation on those cells.
This uses the continuous phase as a training target, so it measures phase information in one state, not trajectory inference.
On the remaining cells, over ten random splits, the $k$-nearest-neighbour readout beat every learned dynamical model on GSE146773 and was clearly better than with permuted labels on GSE121265 (Figure~\ref{fig:fucci}d); ridge regression was worse.
On held-out-bin cells neither readout beat entropic OT.
The U2OS transcriptome therefore carries much more phase information than the iPSC transcriptome.

\paragraph{Limitations.}
Four features of the design limit what the case study shows.
First, coarse FUCCI bins define the snapshots, so FUCCI is not absent from training; only its continuous value is withheld.
Second, the cells cycle independently and are sorted into phase bins afterwards, so the task is to link groups of cells sorted by phase, not snapshots taken at different times.
Third, the OT paths pass through the training cells, so the nearest generated points of a training cell come mostly from its own bin, and the OT readout benefits from the coarse bin labels.
On held-out-bin cells entropic OT still beats the lift--project SDE, but narrowly, and obs-TH or the centroid polygon is best there.
Fourth, each bin is narrower in phase than a real synchronized population, which spreads out over time, so the bins may favor low-noise couplings such as optimal transport.

\end{document}